\documentclass[letterpaper]{article} 
\usepackage{aaai25}  
\usepackage{times}  
\usepackage{helvet}  
\usepackage{courier}  
\usepackage[hyphens]{url}  
\usepackage{graphicx} 
\usepackage{natbib}  
\usepackage{caption} 
\usepackage{algorithm}
\usepackage{algorithmic}

\usepackage{hhline}
\usepackage{amsmath}
\usepackage{amssymb}
\usepackage{amsthm}
\usepackage{lineno}
\usepackage{textcomp}
\usepackage{xcolor}
\usepackage{enumitem}
\usepackage{color}
\usepackage{booktabs}
\usepackage{multirow}
\usepackage{makecell}
\usepackage{colortbl}
\usepackage{scalerel}
\usepackage{bigstrut}
\usepackage{bbding}
\usepackage{etoolbox}

\usepackage{newfloat}
\usepackage{listings}
\DeclareCaptionStyle{ruled}{labelfont=normalfont,labelsep=colon,strut=off} 
\floatstyle{ruled}
\newfloat{listing}{tb}{lst}{}
\floatname{listing}{Listing}

\newlength\savewidth
\newcommand\shline{\noalign{\global\savewidth\arrayrulewidth
                            \global\arrayrulewidth 1.5pt}%
                   \hline
                   \noalign{\global\arrayrulewidth\savewidth}}
\title{AirRadar: Inferring Nationwide Air Quality in China with Deep Neural Networks}
\author{
    Qiongyan Wang\textsuperscript{\rm 1},
    Yutong Xia\textsuperscript{\rm 2},
    Siru Zhong\textsuperscript{\rm 1},
    Weichuang Li\textsuperscript{\rm 1},
    Yuankai Wu\textsuperscript{\rm 3}, \\
    Shifen Cheng\textsuperscript{\rm 5},
    Junbo Zhang\textsuperscript{\rm 4},
    Yu Zheng\textsuperscript{\rm 4},
    Yuxuan Liang\textsuperscript{\rm 1,5}\thanks{Corresponding author. Email: yuxliang@outlook.com}
}

 \affiliations{
    \textsuperscript{\rm 1}The Hong Kong University of Science and Technology (Guangzhou) \\
    \textsuperscript{\rm 2} National University of Singapore \;
    \textsuperscript{\rm 3}Sichuan University \;
    \textsuperscript{\rm 4}JD Intelligent Cities Research \\
    \textsuperscript{\rm 5 }State Key Lab of Resources and Environmental Information System, Chinese Academy of Sciences \\

    \{qiongyanwang, yutong.x, siruzhong, msjunbozhang, msyuzheng, yuxliang\}@outlook.com; \\wli043@connect.hkust-gz.edu.cn; wuyk0@scu.edu.cn; chengsf@lreis.ac.cn 
}

\usepackage{bibentry}

\begin{document}

\maketitle

\begin{abstract}
Monitoring real-time air quality is essential for safeguarding public health and fostering social progress. However, the widespread deployment of air quality monitoring stations is constrained by their significant costs. To address this limitation, we introduce \emph{AirRadar}, a deep neural network designed to accurately infer real-time air quality in locations lacking monitoring stations by utilizing data from existing ones. By leveraging learnable mask tokens, AirRadar reconstructs air quality features in unmonitored regions. Specifically, it operates in two stages: first capturing spatial correlations and then adjusting for distribution shifts. We validate AirRadar's efficacy using a year-long dataset from 1,085 monitoring stations across China, demonstrating its superiority over multiple baselines, even with varying degrees of unobserved data. The source code can be accessed at https://github.com/CityMind-Lab/AirRadar.
\end{abstract}

%

\section{Introduction}

Air pollution poses a significant global health threat, causing numerous premature deaths annually. Pollutants such as CO, SO$2$, and PM${2.5}$ continuously contaminate the air we breathe, exceeding the World Health Organization (WHO) standards for over 90\% of the global population \cite{vallero2014fundamentals}. This widespread problem necessitates urgent and comprehensive solutions for social welfare, aligning with the Sustainable Development Goals \cite{rafaj2018outlook}. 

Given this context, monitoring real-time air quality is crucial for informing effective interventions and policies aimed at safeguarding both public health and the environment. However, the substantial costs associated with establishing monitoring stations, often exceeding 0.2 million USD per station \cite{zheng2013u}, along with their limited quantity and distribution, present a significant challenge to achieving comprehensive data coverage.

\begin{figure}[t]
  \centering
  \includegraphics[width=0.95\linewidth]{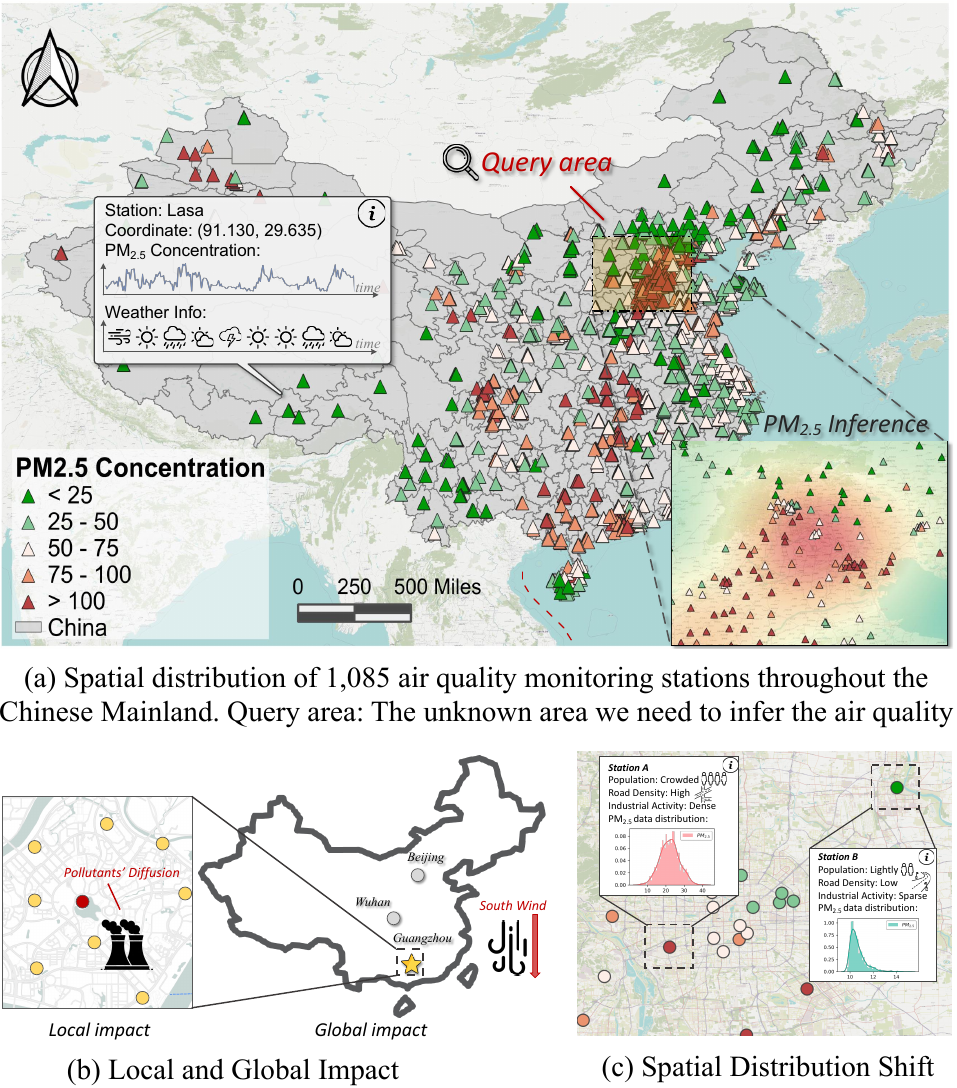}
  \caption{Illustration of air quality inference.} 
  \label{fig:introduction}
\end{figure}

\begin{figure*}[!h]
  \centering
  \includegraphics[width=0.98\textwidth]{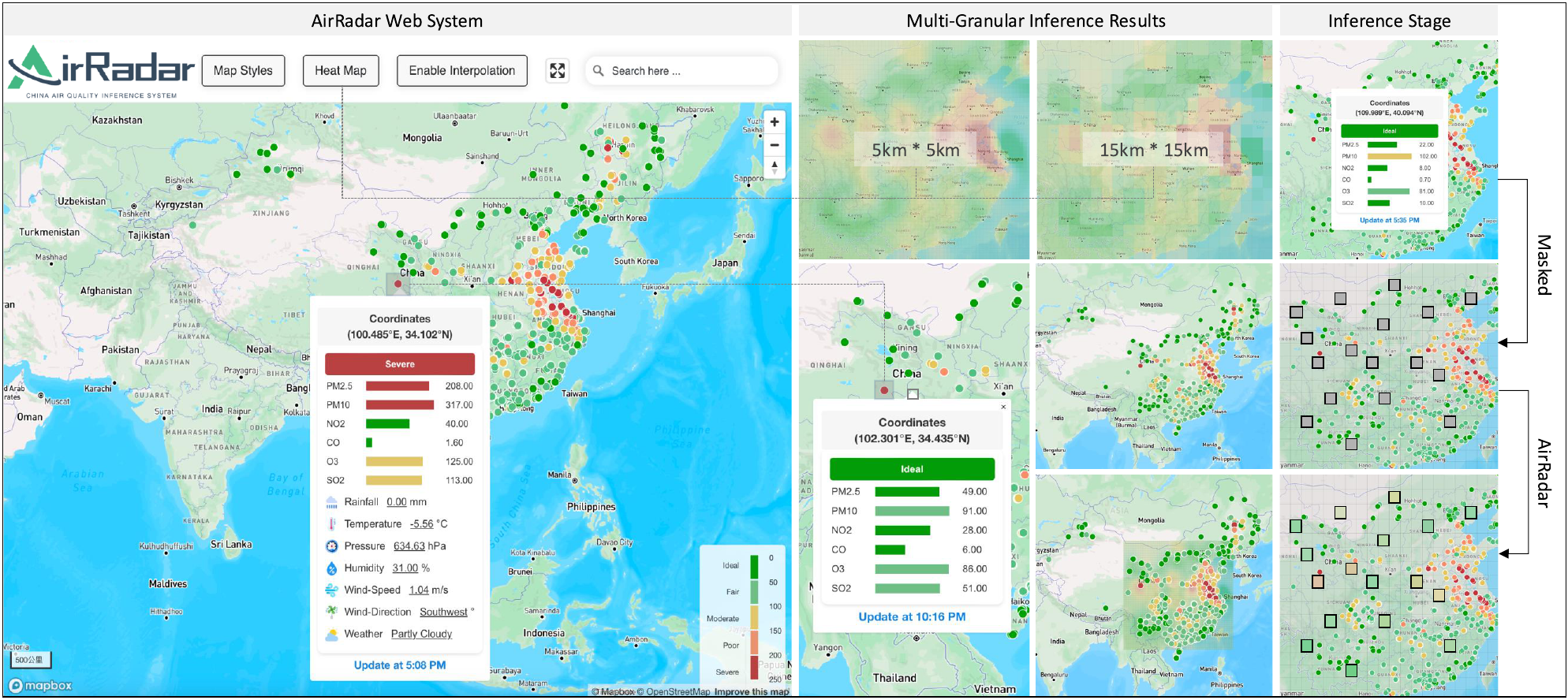}
  \caption{The user interface of our system. It can produce multi-granular inference results.}
  \label{fig:web_system}
\end{figure*}

Over the past decade, \textit{air quality inference} \cite{zheng2013u} has focused on estimating real-time air quality in areas lacking sensors using data from existing ones. Early studies utilized non-parametric methods like k-Nearest Neighbors \cite{cover1968estimation} and Kriging \cite{oliver1990kriging}. With deep learning advances, DL-based model have become prominent for capturing complex spatial correlations among monitoring stations \cite{cheng2018neural,han2021fine,hu2023graph,cheng2024explainable}. However, these models are mostly limited to small-scale tasks at the city level.

In this work, our focus shifts to inferring air quality spanning the \textit{Chinese Mainland} based on thousands of available monitoring stations, as depicted in Figure \ref{fig:introduction}. This encompasses regions with dense sensor coverage as well as remote areas where sensors are sparsely distributed. Such extensive coverage not only delivers more valuable information to the public with high social impacts but also provides a richer dataset that benefits model training \cite{zhao2015multi}. 

Nationwide air quality inference faces notable challenges, particularly in capturing \textit{complex spatial correlations}, aligned with the first law of geography \cite{waters2017tobler}. As illustrated in Figure \ref{fig:introduction}(b), geographically closer stations influence each other more, yet distant sources also affect stations due to pollutant diffusion. Modeling such global spatial relationships via deep GNN layers \cite{kipf2016semi, zhang2024gst_icml} or self-attention mechanisms \cite{liang2023airformer} becomes computationally prohibitive with datasets spanning thousands of stations.

The second challenge arises from \textit{spatial heterogeneity}, aligned with the second law of geography \cite{tobler1970computer}. As shown in Figure \ref{fig:introduction}(c), the concentration of air pollutants is impacted by various region-specific factors, also referred to as context. Unlike city-level inference, nationwide inference must account for diverse regional contexts, making it essential to address spatial heterogeneity for accurate predictions.

To address these issues, we present a new DL-based approach called \textbf{AirRadar}. In unobserved regions -- akin to areas covered in fog -- there exists a critical need for a model that can function like a radar, detecting air quality where direct measurements are unavailable. AirRadar serves this purpose by providing essential air quality data in places lacking direct observation, effectively cutting through the "fog" of uncertainty. To enhance this capability, we employ mask tokens to indicate unobserved data, allowing the model to identify and address gaps in information more accurately.

Targeting the first challenge, we propose a \emph{Spatial Learning Module} that captures spatial correlations locally and globally. The local learner utilizes dartboard projection \cite{liang2023airformer} to effectively process the local information. The global learner uses the  Fast Fourier Transform (FFT) to transform the feature into the frequency domain and capture global relationships with less computational complexity. For the second challenge, our \emph{Causal Learning Module} applies causal theory to interpret causal-effect relationships in real-world contexts, dynamically weighting each context based on its contribution.

\begin{itemize}[leftmargin=*]
    \item We introduce a novel neural network named AirRadar, leveraging both local and global spatial learners to effectively capture complex spatial relationships. AirRadar facilitates air quality inference through masked feature reconstruction, allowing us to utilize a limited dataset to infer air quality in the majority of regions.
    \item We take a causal look and employ backdoor adjustment to address the spatial heterogeneity. Our adaptive weight structure allocates varying weights to the spatial context representations, enabling our model's ability to effectively generalize on data from diverse, unseen spatial contexts.
    \item We evaluate air quality data from thousands of locations under different masking ratios, demonstrating a substantial error reduction ranging from 28.0\%$\sim$44.8\% compared to the state-of-the-art method STFNN. A web-based platform is deployed to verify its practicality (see Figure \ref{fig:web_system}).
\end{itemize}

\begin{figure*}[!t]
  \centering
  \includegraphics[width=0.9\textwidth]{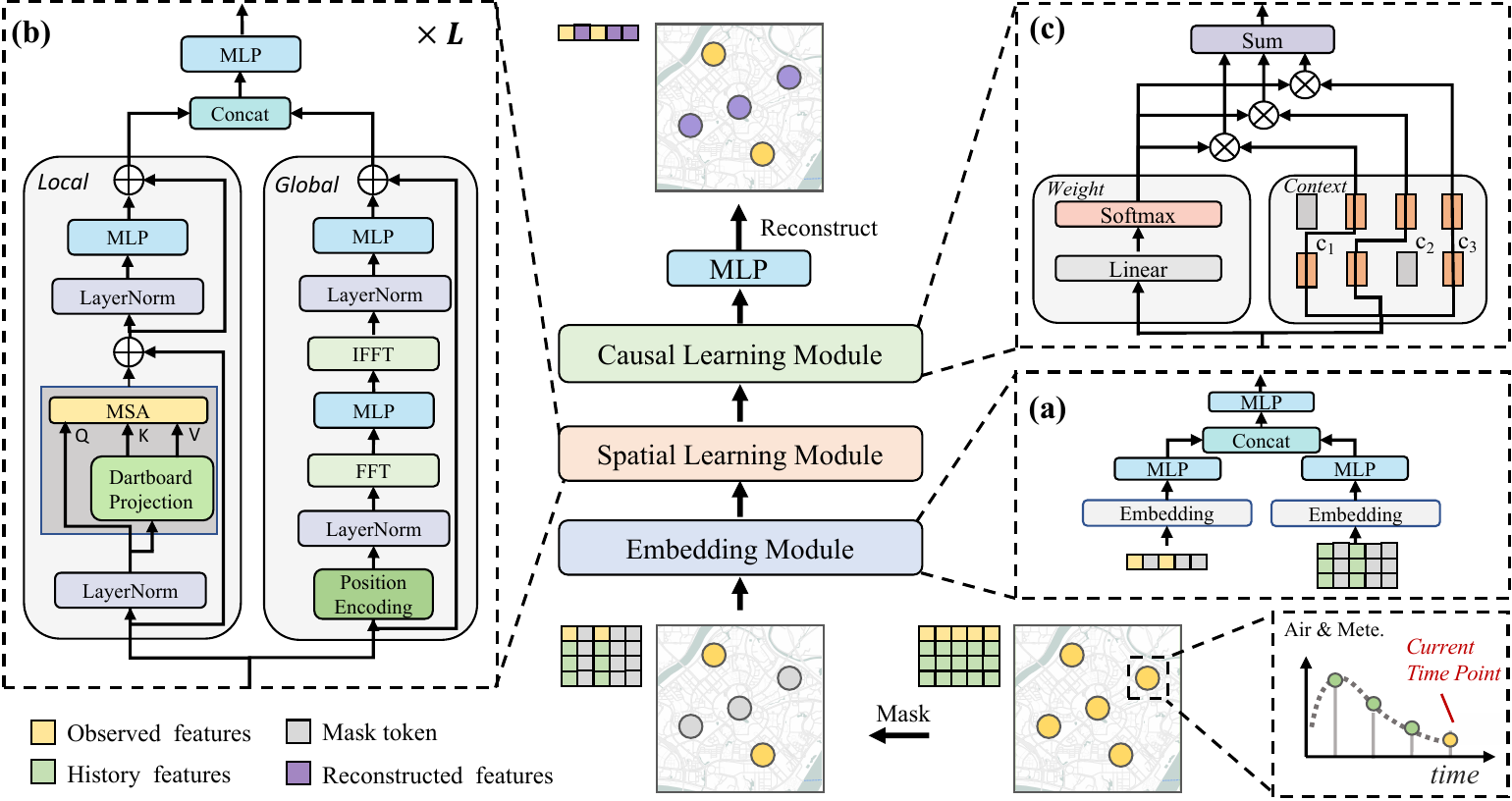}
  \caption{The framework of AirRadar. MLP: multi-layer perception. MSA: multi-head self-attention.} 
  \label{fig:framework}
\end{figure*}

\section{Preliminary}
\subsection{Problem Formulation}
We represent current air quality readings as $\textbf{X} \in \mathbb{R}^{N \times D}$, where $N$ is the number of monitoring stations and $D$ the number of measurements, including air quality and meteorological data. Historical data is denoted as $\textbf{P} \in \mathbb{R}^{N \times T \times D}$, with $T$ representing past time points. A set of nodes $V$ includes observed and unobserved stations, where unobserved nodes lack readings. Given current and historical readings of observed nodes, the goal is to learn a function $f$ to infer air quality at unobserved nodes:
 \begin{equation}
     \mathbf{X}^{'} = f(\Tilde{\textbf{X}},\Tilde{\textbf{P}})
 \end{equation}
where $\sim $ represents masking. $\textbf{X}^{'}$ denotes the inferred results for all nodes, with PM$_{2.5}$ as the primary pollutant.

\subsection{Related Works}
\noindent \textbf{Air Quality Inference.} Accurately inferring air quality is paramount for societal well-being. As air quality research progresses, models integrate multi-source data to make detailed inferences. Traditional methods \cite{fawagreh2014random, cheng2020lightweight, cheng2024explainable} often rely on distance-based features, limiting their ability to capture site significance. Different from these models, certain studies \cite{cheng2018neural,han2021fine} incorporate the attention mechanism to assign sensor weights automatically but face challenges with sparse data and providing detailed insights. Recently, STGNP \cite{hu2023graph} introduced the Graph Neural Process for spatial extrapolation tasks. STFNN \cite{ijcai2024p803} integrates both spatial and temporal viewpoints, utilizing fields and graphs to enhance the accuracy of air quality inference. While these models have made strides in air quality inference, they fall short in addressing two critical challenges: the \textit{complex spatial correlations} and \textit{spatial heterogeneity}. 

\noindent \textbf{Neural Networks for Spatial Learning.} Neural networks have been widely applied in spatial learning, leveraging various architectures to capture spatial dependencies. GNNs \cite{wu2022graph,li2022cell} are commonly used to model spatial relationships in non-Euclidean space. CNNs \cite{guo2019deep,khosravi2020convolutional} are often employed to capture local spatial features by processing data with grid-like topologies, such as images or regular grids. Recently, Transformers \cite{wu2020hierarchically,hao2024urbanvlp} have also been adapted for spatial learning, offering an attention mechanism to capture long-range dependencies. Each of these approaches contributes uniquely to enhancing spatial learning capabilities in different tasks. 

\noindent \textbf{Causal Inference.} Causal inference \cite{pearl2016causal} aims to explore causal relationships between variables, promoting stable and robust learning and inference processes. The integration of deep learning techniques with causal inference has demonstrated significant potential in spatial-temporal mining tasks \cite{xia2024deciphering,ijcai2024p248}. In this study, we apply causal methods to address spatial heterogeneity in the air quality inference task.

\section{Methodology}
Figure \ref{fig:framework} depicts the framework of AirRadar, which consists of input masking and three learning modules. To address the issue of the target node lacking prior information, a shared learnable token is employed to mask the features of target nodes. During training, stations are masked as target nodes, while areas lacking stations serve as target nodes during inference (See Figure \ref{fig:web_system} Right). We subsequently process the masked inputs with the following modules:
\begin{itemize}[leftmargin=*]
    \item \emph{Embedding module}: We utilize a fully-connected layer to transform the masked air quality data into feature space and integrate observed features with historical features to obtain an initial representation for each node.

    \item  \emph{Spatial Learning Module}: This module integrates spatial correlation from both local and global perspectives. The local part uses dartboard projection to efficiently incorporate local spatial information, while the global part transforms the feature from the spatial domain to the frequency domain to effectively capture global relationships.

    \item  \emph{Causal Learning Module}: 
    We begin by constructing a structural causal model to illustrate how various contexts influence air quality. We then implement backdoor adjustment by dynamically allocating weights to each context, effectively mitigating confounding effects.

\end{itemize}

\subsection{Embedding Module}
 Considering that observed data and historical data can provide essential information for inferring the air quality of unobserved nodes, we have taken these data into our input. 
 However, the target nodes are without any prior information. Inspired by \texttt{MAE} \cite{he2022masked}, we employ a shared learnable token $\textbf{o}$ to mask the data of target nodes. Then we transform the masked data into feature space, resulting in $\hat{\mathbf{X}} \in \mathbb{R}^{N\times E}$ and $\hat{\mathbf{P}} \in \mathbb{R}^{N \times  E}$. Finally, we combine the observation and historical data to create an initial spatial representation for each node, denoted as $\mathbf{H} \in \mathbb{R}^{N \times 2E}$. 

\subsection{Spatial Learning Module}
We introduce the Spatial Learning Module, a novel spatial learner designed to capture spatial correlations from local and global perspectives based on the principle of spatial auto-correlation. 
Our Spatial Learning Module boasts a substantial receptive field with computational complexity less than quadratic. Figure \ref{fig:framework}b outlines its pipeline. Each block (indexed by $l$) takes the hidden state ${\textbf{Z}}^{l-1}$ as input, with $\mathbf{H}$ serving as the initial embedding $\textbf{Z}^{0}$. This input is split into two parts: the local module (left side) and the global module (right side).  In the subsequent parts of this section, we will detail these two key parts respectively.

\subsubsection{Local Spatial Correlation.} 
\begin{figure}[!t]
  \centering
  \includegraphics[width=\linewidth]{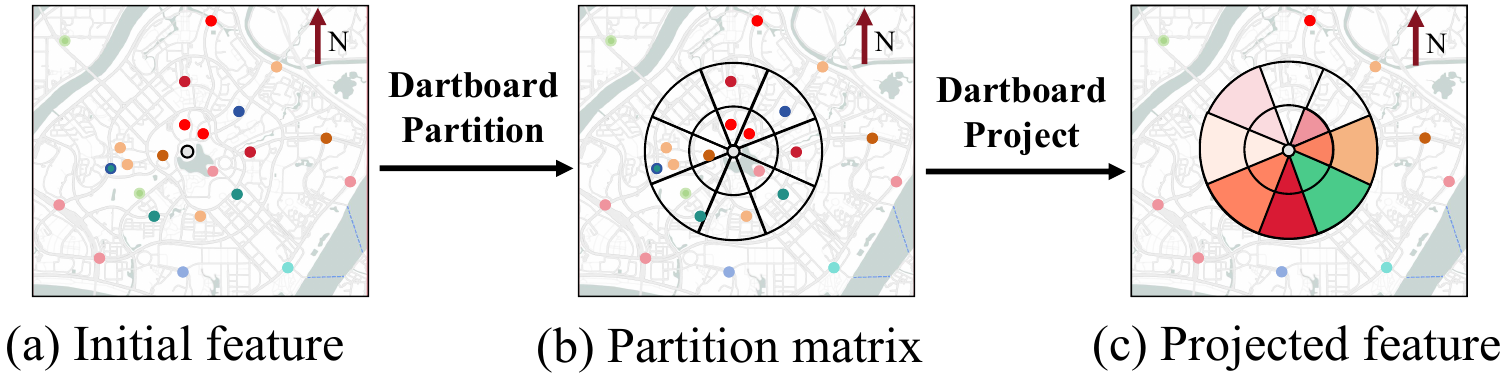}
  \caption{Dartboard projection: For a query node (gray point), its local range is partitioned into regions by circles and lines. The central angle aligns with wind directions.}
  \label{fig:Dartboard}
\end{figure}

Beyond local emissions, a place's air quality is also affected by neighboring areas. Directly using standard MSA will result in a square increase in costs, thus our objective is find a new way to \textit{efficiently} capture local spatial correlations. We drew inspiration from \cite{liang2023airformer} and used a dartboard projection to achieve this. As depicted in Figure~\ref{fig:Dartboard}, node $i$'s local range is defined using a dartboard, which divides its neighborhood space into several distinct regions. Nodes positioned beyond the outermost circle fall outside the local range. Different from \cite{liang2023airformer}, we modify the orientation of the central angle to align with the potential wind direction, thereby enhancing integration with external factors such as wind direction and the interpretability of the model. 

The latent features $\textbf{Z}^{l-1}$ are used to generate the query vector $\textbf{Q}_{h}$ for each node. For the query nodes, the key vector and value vector are obtained via dartboard projection. The dartboard projection matrix for each node $i$ is represented as $\textbf{M}_{i} \in \mathbb{R}^{G \times N}$, indicating how nearby nodes are mapped to $G$ regions defined by the line fragments and circles. Here, each entry $\textbf{m}_{js}=1$ signifies that the $s$-th node belongs to the $j$-th region. We employ the dartboard projection matrix to project the node features into local regional representations. The projected features of each region $\textbf{R}_{i} \in \mathbb{R}^{G\times 2E}$ are generated by:
\begin{equation}
    \textbf{R}_{i} = \textbf{M}_{i}\textbf{F} = \sum\nolimits_{\forall j}\textbf{m}_{ij}\textbf{f}_{j},
\end{equation}
with projection matrix $\textbf{M}_{i}$ and normalized node features $\textbf{f}_{j} \in \mathbb{R}^{2E}$.
The ensuing step involves concatenating these representations after dartboard projection $\textbf{R}=[\textbf{R}_{1}, \textbf{R}_{2},\ldots, \textbf{R}_{N}]$. 
Subsequently, we feed this regional representation into linear layers to generate the key vector $\textbf{K}_{h} \in \mathbb{R}^{N \times G\times 2E}$ and value vector $\textbf{V}_{h} \in \mathbb{R}^{N \times G\times 2E}$. The implementation of multi-head self-attention efficiently captures detailed local spatial correlations between the query node and its neighboring regions, as articulated by:
\begin{equation}
    \textbf{X}_{h} = \operatorname{Softmax}(\alpha \textbf{Q}_{h}\textbf{K}_{h}^{T}+\textbf{B}_{h})\textbf{V}_{h},
\end{equation}
where $\textbf{B}_{h} \in \mathbb{R}^{N \times G}$ is a learnable relative position encoding.

This local spatial learner integrates domain knowledge of air diffusion while addressing computational expense. With a smaller number of regions ($G \ll N$), the computational complexity, $O(NGE)$, grows linearly with the number of stations, proving more efficient than standard MSA  with complexity $O(N^{2}E)$.

\subsubsection{Global Spatial Correlation.} In the context of air quality inference, beyond accounting for local spatial relationships, it is essential to incorporate considerations of global spatial relationships, i.e., impact from other cities. 

To avoid the computational complexity growth with the quadratic cost, we consider using the Fourier Neural Operator (FNO) to reduce the computational burden, inspired by \cite{li2020fourier}. We first conceptualize the process of self-attention as a continuous kernel integrals \cite{guibas2021adaptive,li2020neural}, denoted as follows:
\begin{equation}
    \zeta(X)(i) = \int_{D}\kappa(i,j)X(j)\ dj \quad \forall i\in \mathbb{D},
\end{equation}
where $\zeta(X)(i)$ is equivalent to the $i$-th row of the similarity matrix $Att \in \mathbb{R}^{N \times N}$. 

This means that when calculating $\zeta(X)(i)$, the information on the entire domain is considered and the global information is captured. Furthermore, this operation can be efficiently implemented by FFT with computational complexity $O(NElogN)$:
\begin{equation}
    \zeta(X)(i) = F^{-1}(F(\kappa)\cdot F(X))(i) \quad i \in \mathbb{D},
\end{equation}
where $\cdot$ denotes matrix multiplication and $F$, $F^{-1}$ denote the FFT and its inverse, respectively. 

We treat each node as the token in the sequence. For the input feature $\textbf{Z}^{l-1} \in \mathbb{R}^{N \times 2E}$, we need to define the complex-valued weight tensor $\hat{\textbf{W}}$ to parameterize the kernel.  To reduce the parameter count, we impose a block diagonal structure on $\hat{\textbf{W}}$, where it is divided into $\hat{K}$ weight blocks of size $2E/\hat{K} \times 2E/\hat{K}$. Therefore, the global spatial mixing operations are defined as follows:
\begin{equation}
\begin{aligned}
    &\mathbf{g}_{i} = [\operatorname{FFT}(\mathbf{Z}^{l-1})]_{i} \\ 
    &\Tilde{\mathbf{g}}^{k}_{i} = \hat{\textbf{W}}_{k}\mathbf{g}^{k}_{i} + \textbf{b}_{k} \quad k = 1,2,\cdots,\hat{K} \\
    &\hat{\mathbf{g}}_{i} = [\operatorname{IFFT}(\operatorname{S}(\Tilde{\mathbf{g}},\lambda))]_{i}
\end{aligned}
\end{equation}
where $\hat{\textbf{W}}$ are shared for all nodes, the computational complexity of node mixing is $O(NE^{2}/ \hat{K})$. The scalar $\lambda$ represents the sparsity threshold, while $\operatorname{S}$ denotes the soft-thresholding and shrinkage operation.

To efficiently capture global spatial relationships, we transform input features from spatial to frequency domains and
employ shared weights for spatial mixing. The computational complexity of this FNO with shared weights is $O\left(NE^{2} / \hat{K} + NE \log N\right)$, more efficient than $O(N^{2}E + 3NE^{2})$, especially with increasing node count. After obtaining local and global spatial representations for each node, we combined them, denoted as $\textbf{Z}^{l} \in \mathbb{R}^{N \times 2E}$.

\subsection{Causal Learning Module}
\begin{figure}[!h]
    \centering
    \includegraphics[width=0.5\linewidth]{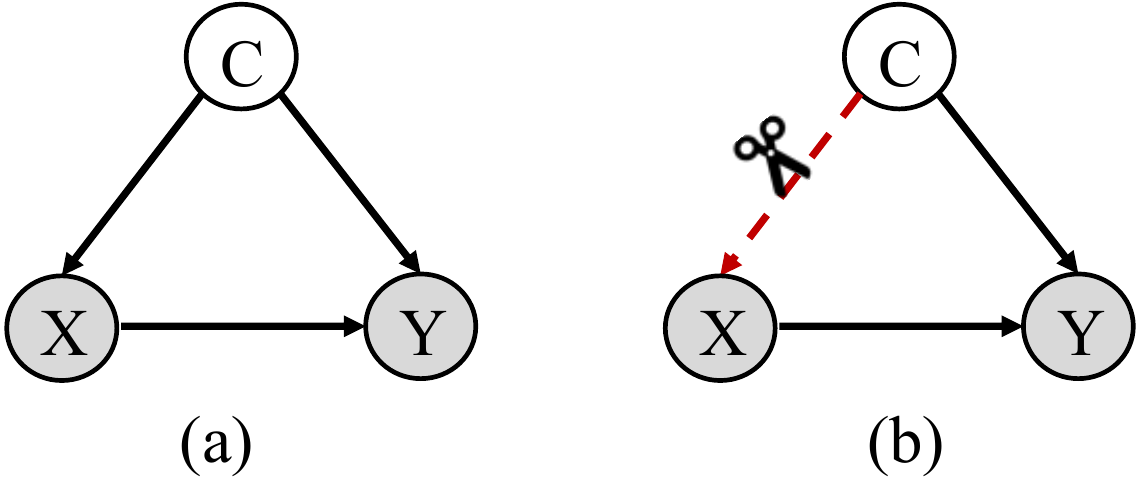}
    \caption{SCMs of (a) Data generation under Real-world Scenarios; (b) Backdoor Adjustment for $C$.}
    \label{fig:SCM}
\end{figure}

In air quality inference, as the geographical scope widens, the challenge of spatial heterogeneity emerges. To tackle this, we leverage causality theory by establishing a Structural Causal Model (SCM) \cite{pearl2000models}. In the SCM shown in Figure \ref{fig:SCM}a, $C$, $X$, and $Y$ denote random variables for spatial context, observed data, and target area air pollutants, respectively, with arrows representing causal-effect relationships. Based on the above definitions, the causal relationships in Figure \ref{fig:SCM}a can be denoted as
$P(X,Y|C) = P(X|C)P(Y |X,C)$. 
\subsubsection{Confounder \& Intervention.} We observe that there exists a backdoor path $X \leftarrow C \rightarrow Y$ between $X$ and $Y$, where context $C$ acts as a confounding factor. However, this confounding influence of $C$ is undesirable for accurate inference, as it introduces potential bias in the model's inference. 

The \textit{backdoor adjustment} \cite{yang2022towards,sui2022causal,xia2024deciphering} can cut off the causal relationship from $C$ to $X$ to mitigate the confounding effects, as shown in the red dashed arrow in Figure \ref{fig:SCM}b. Therefore, we employ a backdoor adjustment to estimate $Y$ via $P(Y|do(X))$ by stratifying into discrete pieces $C = \{c_{i}\}_{i=1}^{|C|}$ as:
\begin{equation}
    P(Y|do(X)) = \sum\nolimits_{i=1}^{|C|}P(Y|X,C=c_{i})P(C=c_{i})
    \label{eq:backdoor}
\end{equation}
Although the backdoor path may seem effectively severed, optimizing with Eq.~\ref{eq:backdoor} becomes intractable due to challenges in defining $c_{i}$ and the unknown prior distribution $P(C)$. To overcome this, we treat $C$ as a latent variable obtained by a variational distribution $Q(C|X)$ \cite{yang2022towards}.

\begin{figure}
    \centering
    \includegraphics[width=0.8\linewidth]{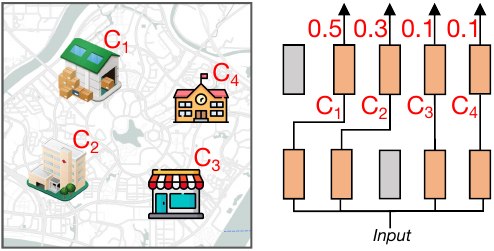}
    \caption{An explanation of how AirRadar combines the influence of each context. $c_{k}$ denotes the context type.}
    \label{fig:context}
\end{figure}

Estimating this distribution is facilitated by the Causal Learning Module, illustrated in Figure \ref{fig:framework}c. Assuming that the contexts are independent can lead to overfitting issues, thus we denote a context as a 0-1 matrix, i.e.,
\begin{equation}
    c_{k} = stack[c_{k}^{1},c_{k}^{2},... ,c_{k}^{U}]
\end{equation}
where $U$ denotes the number of layers and $K$ denotes the number of contexts. Each context type corresponds to a specific combination of inference units. To comprehensively consider the influence of various contexts, we design a network that can dynamically allocate weights:
\begin{equation}
    \mathbf{Y}^{l+1} = \sum_{k=1}^{K} \Psi^{l}(\mathbf{Y}^{l})[k] \, \Phi_k^l\left(\mathbf{Y}^l \right),
\end{equation}
where $\Psi$ is the dynamic weight allocation structure and $\Phi$ is the context-specific encoder. The computation represents a realization of $c$ drawn from the conditional distribution $Q(C|X = \textbf{\textit{X}})$. As shown in Figure \ref{fig:context}, different contexts may be linked by sharing the same inference units at certain layers, serving as an effective inductive bias that helps the model eliminate the influence of spatial heterogeneity and generalize to unseen areas.

\subsection{Inference \& Optimization}
Upon obtaining the final feature representation \textbf{Y}, we employ an MLP for feature reconstruction to restore the air quality data of all nodes as:
\begin{equation}
    \mathbf{X}^{'} = \operatorname{MLP}(\mathbf{Y})
\end{equation}
where $\textbf{X}^{'} \in \mathbb{R}^{N \times D}$ is the reconstrued data for all nodes. This process culminates in the completion of the target inference. For optimization, we utilize the L1 loss as the loss function $\mathcal{L}$, expressed as follows:
\begin{equation}
\mathcal{L}(\mathbf{X}, \mathbf{X}^{'}) = \frac{1}{N_{u}} \sum\nolimits_{v_{i} \in \Tilde{V}} \lvert \mathbf{X}_{i} - \mathbf{X}^{'}_{i} \rvert
\end{equation}
where $N_{u}$ is the number of unobserved nodes. Note that we exclusively compute the loss for the target nodes. 

\begin{table*}[!ht]
\centering
\footnotesize
\renewcommand{\arraystretch}{1.3}
\setlength{\tabcolsep}{4.5pt}
\caption{Model comparison on the nationwide dataset. The symbol $\Delta$ represents the reduction in MAE compared to the state-of-the-art method STFNN \protect\cite{ijcai2024p803} The missing ratio represents the proportion of unobserved nodes to all nodes.}
\begin{tabular}{c|c|cccc|cccc|cccc}
\shline
\multirow{2}{*}{\textbf{Model}} &\multirow{2}{*}{\textbf{Year}} & \multicolumn{4}{c|}{\textbf{Missing Ratio = 25\%}} & \multicolumn{4}{c|}{\textbf{Missing Ratio = 50\%}} & \multicolumn{4}{c}{\textbf{Missing Ratio = 75\%}} \\
\cline{3-14}
& & \textbf{MAE} & \textbf{$\boldsymbol{\Delta}$} & \textbf{RMSE} & \textbf{MAPE} & \textbf{MAE} & \textbf{$\boldsymbol{\Delta}$} & \textbf{RMSE} & \textbf{MAPE} & \textbf{MAE} & \textbf{$\boldsymbol{\Delta}$} & \textbf{RMSE} & \textbf{MAPE} \\
\hline
KNN &1967 &30.50  & +173.8\% &65.40  &1.36  &30.25  &+167.2\%  &72.23  &0.71  & 34.07 & +202.3\% & 74.55 & 0.64 \\
RF&2001 &29.22  & +162.3\% &68.95  &0.76  &29.71  & +162.5\% &71.61  &0.75  & 29.82 & +164.6\% & 70.99 & 0.74 \\ \hline
SGNP &2019 &23.60 &+96.0\% &37.58 &0.83 & 24.06 & +112.5\% & 37.08 & 0.93 & 21.68 & +89.1\% & 33.68 & 0.84 \\
STGNP &2022 & 23.21 & +90.6\% & 38.13 & 0.62 & 21.95 & +93.9\% & 37.13 & 0.67 & 19.58 & +73.7\% & 31.95 & 0.69 \\ \hline
VAE &2013 &28.49 & +155.7\% & 67.11 & 0.94 & 28.92 & +155.5\% & 69.67 & 0.94 & 29.00 & +157.3\% & 69.11 & 0.93 \\
GAE &2016 &12.63 & +13.4\% & 23.80 & 0.46 & 12.78 & +12.9\% & 24.11 & 0.46 & 12.57 & +11.5\% & 23.73 & 0.46 \\
GraphMAE&2022 &12.40 & +11.3\% & 23.20 & 0.46 & 12.32 & +8.8\% & 23.11 & 0.46 & 11.59 & +2.8\% & 21.51 & 0.43 \\ \hline
MCAM &2021 &23.94 & +120.9\% & 36.25 & 0.95 & 25.01 & +103.0\% & 37.94 & 0.92 & 25.19 & +123.5\% & 37.82 & 1.04 \\
STFNN&2024 &11.14 & - & 23.20 & 0.39 & 11.32 & - & 19.91 & 0.41 & 11.27 & - & 19.86 & 0.41 \\ \hline
\rowcolor{gray!20} AirRadar&- & \textbf{6.41} & \textbf{-40.0\%} & \textbf{12.60} & \textbf{0.24} & \textbf{6.79} & \textbf{-44.9\%} & \textbf{12.90} & \textbf{0.26} & \textbf{8.11} & \textbf{-28.0\%} & \textbf{14.84} & \textbf{0.29} \\
\shline
\end{tabular}
\label{tab:results}
\end{table*}

\section{Experiments}
In this section, we conduct experiments to address the following research questions:
\begin{itemize}[leftmargin=*]
    \item \textbf{RQ1}: Can AirRadar outperform existing approaches on real-time air quality inference in the Chinese Mainland?
    \item \textbf{RQ2}: How does each component affect performance?
    \item \textbf{RQ3}: What is the effect of the major hyperparameters?
    \item \textbf{RQ4}: How is the practicality of AirRadar?
\end{itemize}

\subsection{Experimental Settings}
\subsubsection{Datasets.} Our system (currently anonymous) collected nationwide air quality datasets for a year (Jan. 1, 2018 to Dec. 31, 2018). Figure \ref{fig:introduction} illustrates the distribution of monitoring stations, showcasing broad spatial coverage exceeding previous datasets. Meteorological data was also incorporated to consider exogenous influences. Input features encompass continuous (e.g., pollutant concentrations, temperature) and categorical (e.g., weather conditions, wind direction) variables. Categorical features were transformed into real-valued vectors via embedding. Following precedents \cite{cheng2018neural,hu2023graph}, known air quality data was utilized to infer PM$_{2.5}$ values for other sites. The dataset was split into training (60\%), validation (20\%), and testing (20\%) sets, comprising 5,256, 1,752, and 1,752 instances respectively.

\subsubsection{Implement Details.} Our model, implemented with PyTorch 2.1.0 on an NVIDIA RTX 2080Ti GPU, uses the Adam optimizer with a batch size 32, and an initial learning rate of $5 \times 10^{-3}$. The hidden size ($E$) is 32. We randomly mask a percentage of node features and include 2 spatial learning blocks. The Local part has 2 heads with 50km and 200km ranges. The global part comprises 2 weight blocks, and the Causal Learning Module uses 4 contexts. 

\subsubsection{Baselines \& Evaluation Metrics.}
We compare our AirRadar with strong baselines that belong to four categories:
\begin{itemize}[leftmargin=*]
    \item \textbf{Statistical models}: \textbf{KNN} \cite{cover1968estimation} and \textbf{RF} \cite{fawagreh2014random} are conventional baselines for spatial interpolation, e.g., air quality inference.
    \item \textbf{NP-based models}: We take several popular NP-based models as baselines, i.e., \textbf{SGNP} which is a modification of \cite{singh2019sequential}, and \textbf{STGNP} \cite{hu2023graph}.
    \item \textbf{Autoencoder-based models}: We also take some popular Autoencoders as baselines, i.e., \textbf{VAE} \cite{kingma2013auto}, \textbf{GAE} and \textbf{GraphMAE} 
    \cite{hou2022graphmae}.
    \item \textbf{Other models}: We select a couple of other models for comparison, including \textbf{MCAM} \cite{han2021fine}, \textbf{STFNN} \cite{ijcai2024p803}.
\end{itemize}
To evaluate our model, we follow STFNN \cite{ijcai2024p803} to employ Mean Absolute Error (MAE), Root Mean Square Error (RMSE), and Mean Absolute Percentage Error (MAPE) as the evaluation metrics. 

\subsection{Model Comparison (RQ1)}
To answer RQ1, we perform a model comparison in terms of MAE, RMSE, and MAPE. Table \ref{tab:results} presents the experimental results over the nationwide dataset. Our model significantly outperforms all competing baselines over all kinds of missing ratios. Compared to the STFNN, our approach reduces MAE by approximately 28\% to 44.9\% in each missing ratio. There are three primary reasons for the efficacy of our model. Firstly, our method effectively captures comprehensive spatial correlations. Secondly, AirRadar leverages backdoor adjustment to eliminate confounding effects and dynamically assigns weights to each potential context representation. Lastly, using the learnable token to represent the target node empowers robust inference even for unobserved nodes. In contrast, traditional models (KNN, RF) lack strong learning capabilities, while AE-based, NN-based, and NP-based models struggle with spatial range and diversity, limiting their adaptability for nationwide air quality inference.

\subsection{Ablation Study (RQ2)}
\begin{figure}[!h]
  \centering
  \includegraphics[width=0.48\textwidth]{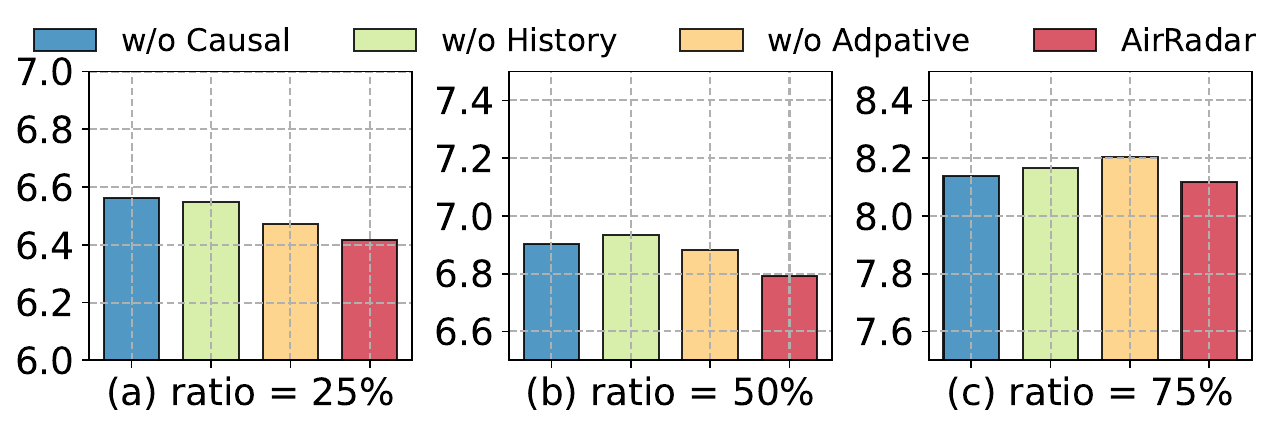}
  \caption{Effects of modules across different missing ratios.}
  \label{fig:core_components}
\end{figure}

To evaluate the individual contributions of each component to our model's overall performance and address RQ2, we conduct ablation studies. The results are presented in Figure \ref{fig:core_components} and Table \ref{tab:DA-integrator}. Each study involves modifying only the corresponding part while keeping other settings unchanged.

\noindent \textbf{Effects of Historical Data}. As a pivotal component of AirRadar, historical data plays a fundamental role in offering prolonged observations of the environmental background, encompassing factors such as meteorological conditions and seasonal variations. As illustrated in Figure \ref{fig:core_components}, the exclusion of historical data leads to a noticeable decline in performance across all missing ratios.

\noindent \textbf{Effects of Spatial Learning Module}. To assess the impact of the Spatial Learning Module, we examine the following variants for comparison: a) \textbf{w/o Local}: The local component in the Spatial Learning Module is disabled, meaning that no local spatial relationships are considered. b) \textbf{w/o Global}: The global component in the Spatial Learning Module is omitted, indicating that the global spatial correlation is not taken into account in this model. The results are shown in the Table \ref{tab:DA-integrator}. When eliminating both local and global components, we observe a significant deterioration in MAE, especially from removing the global part. This highlights the importance of considering spatial correlations from both perspectives. Second, AirRadar obtains lower inference errors overall missing ratios, while running \textbf{44\%} faster than GraphMAE. This advantage indicates that our Spatial Learning Module holds significant promise as a fundamental component for capturing complex spatial dependencies.

\begin{table}[!h]
\tabcolsep=2.8mm 
\footnotesize
\centering
\caption{Effects of Spatial Learning Module under different missing ratios. Time/epoch: seconds per training epoch.}
\begin{tabular}{l||c|c|c|c}
\hline 
\textbf{Variant} & \textbf{Time/epoch} & \textbf{25\%}          & \textbf{50\%}          & \textbf{75\%}          \\ \hline
\hline
w/o local        & 27       & 8.67          & 7.45          & 9.09          \\ 
w/o global       & 24       & 8.43          & 9.22          & 11.68         \\ 
GraphMAE         & 74       & 12.50         & 12.43         & 12.92         \\ \rowcolor{gray!20}
AirRadar         & 41       & \textbf{6.41} & \textbf{6.79} & \textbf{8.11} \\ \hline
\end{tabular}
\label{tab:DA-integrator}
\end{table}

\noindent \textbf{Effects of Causal Learning Module}. To evaluate the effectiveness of the Causal Learning Module in addressing spatial heterogeneity challenges, we compare our model with the e following variants: a) \textbf{w/o Causal}: The entire Causal Learning Module is omitted from our AirRadar. b) \textbf{w/o Adaptive}: The dynamic weight allocation mechanism is excluded from the Causal Learning Module. The outcomes are presented in Figure \ref{fig:core_components}. When the causal module is omitted, the model fails to mitigate the influence of potential context and, consequently, cannot address the spatial heterogeneity issue. Absent the adaptive module, the model is unable to account for varying impacts introduced by different contexts, resulting in a performance decline.

\subsection{Hyperarmeter Study (RQ3)}
\begin{figure}[!h]
  \centering
  \includegraphics[width=0.47\textwidth]{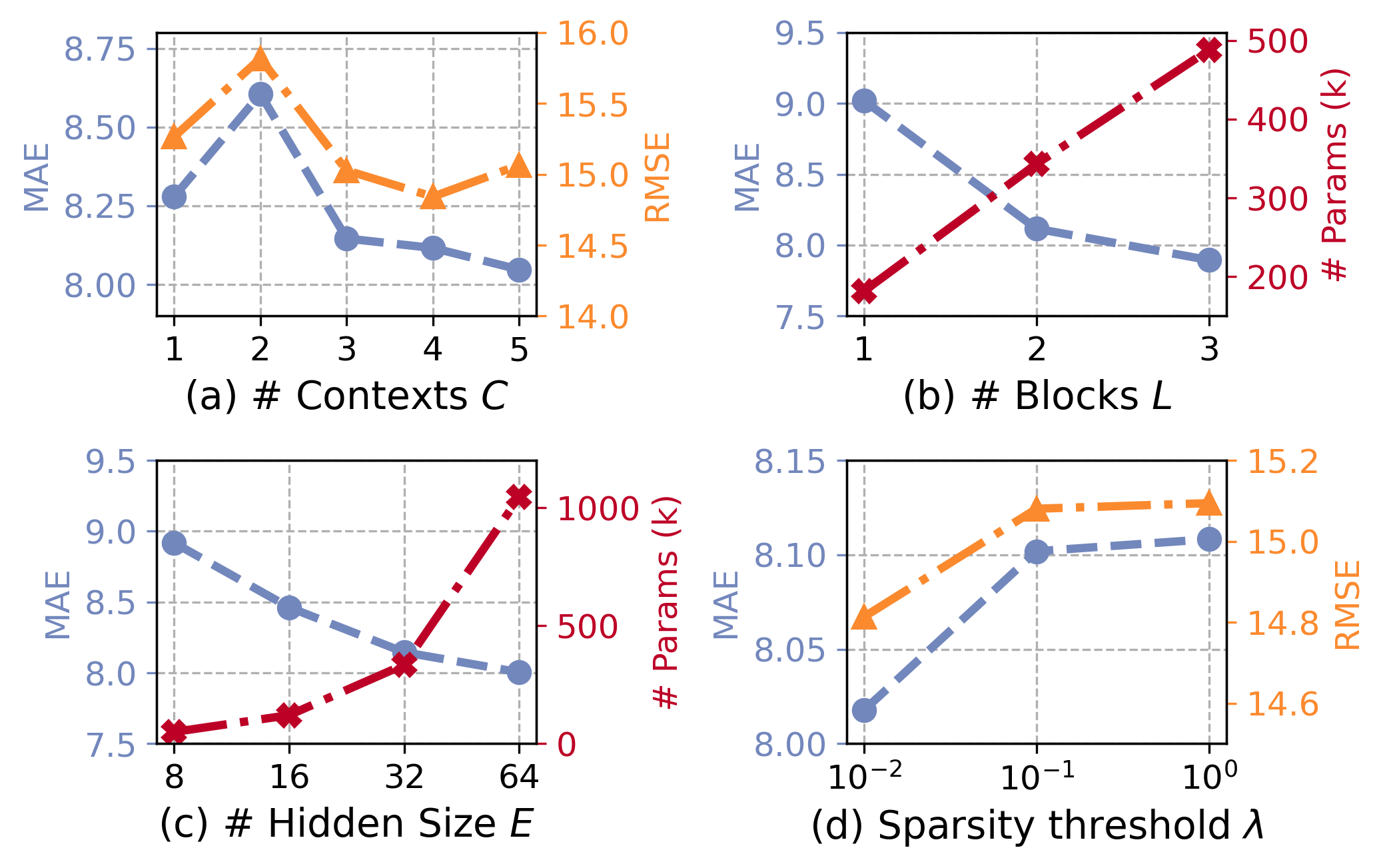}
  \caption{Hyperparameter study when missing ratio=75\%.}
  \label{fig:hp}
\end{figure}

 In response to the final research question, we explore the impact of varying the hidden dimension $E$, the number of Spatial Learning blocks $L$, the number of contexts $C$, and the sparsity threshold $\lambda$ on the MAE, with a fixed missing rate of 75\%. Figure \ref{fig:hp}a presents the results of AirRadar with different values of $C$. Interestingly, we observe that the performance is not strictly proportional to the number of contexts, and we establish $C = 4$ as our default setting based on these findings. We experiment with different sparsity thresholds $\lambda$ ranging from 0 to 1. The MAE reaches its peak at $\lambda=10^{-2}$, demonstrating the efficacy of the sparsity measure.

Figure \ref{fig:hp}b investigates the number of Spatial Learning blocks ($L$). It becomes evident that the model's performance improves as $L$ increases, but this improvement comes at the cost of a rapid growth in the number of parameters. Notably, when $L = 3$, we achieve a similar level of accuracy without the excessive parameter increase. Figure \ref{fig:hp}c further delves into the influence of the hidden dimension $E$. Here, we make two key observations: 1) A very small $E$ (e.g., 8) results in diminished performance due to its limited capacity. 2) $E = 64$ yields similar performance, but it comes with a substantial increase in computational cost.

\subsection{Practicality (RQ4)}\label{subsec:RQ4}
To verify the model's practicality, we deploy a web-based system for monitoring air quality and inferring nationwide air quality in China, as shown in Figure \ref{fig:web_system}. Each point on the map corresponds to an air quality monitoring station and the
the associated number signifies the respective air quality data at that station. The system's practicality lies in integrating real-time data from numerous stations, providing accessible air quality information nationwide. 

\section{Conclusion \& Future Work}
In this paper, we introduce AirRadar for nationwide air quality inference in China. AirRadar efficiently captures intricate spatial correlations from both local and global perspectives, addressing the challenge of spatial heterogeneity through backdoor adjustment techniques with an adaptive weight mechanism. Our model can achieve state-of-the-art performance on a nationwide air quality dataset with different missing ratios. Accompanying our model, we develop a web system capable of leveraging air quality data from existing monitoring stations to infer air pollutants of regions without monitoring stations. In the future, we will consider online learning and explore the application of our model in other domains, such as traffic and climate. 

\section{Acknowledgments}
This work is supported a grant from State Key Laboratory of Resources and Environmental Information System. This work is also supported by the National Natural Science Foundation of China (No. 62402414), the Guangzhou-HKUST(GZ) Joint Funding Program (No. 2024A03J0620), Guangzhou Municipal Science and Technology Project (No. 2023A03J0011),  the Guangzhou Industrial Information and Intelligent Key Laboratory Project (No. 2024A03J0628), and Guangdong Provincial Key Lab of Integrated Communication, Sensing and Computation for Ubiquitous Internet of Things (No. 2023B1212010007).

\bibliography{aaai25}

\end{document}